\documentclass{bmvc2k}

\usepackage{times}
\usepackage{epsfig}
\usepackage{graphicx}
\usepackage{amsmath}
\usepackage{amssymb}
\usepackage{booktabs}
\usepackage{mathtools}
\usepackage{layouts}
\usepackage{multirow}
\usepackage{enumitem}
\usepackage{color}

\newcommand{\keypoint}[1]{\vspace{0.1cm}\noindent\textbf{#1}\quad}
\newcommand{\cut}[1]{}
\usepackage{algorithm}
\usepackage{algpseudocode}

\usepackage[nopar]{lipsum}
\usepackage{arydshln}
\DeclareMathAlphabet\mathbfcal{OMS}{cmsy}{b}{n}

\usepackage{array}
\usepackage{colortbl}
\usepackage{hhline}
\usepackage{soul}
\usepackage{pgfplots}
\usepackage{wrapfig}
\usepackage{pgf}
\usepackage{stmaryrd}
\pgfplotsset{width=10cm,compat=1.9}
\definecolor{LightCyan}{rgb}{0.95,0.95,0.95}
\newcolumntype{a}{>{\columncolor{LightCyan}}c}

\title{Towards Generative Class Prompt Learning for Fine-grained Visual Recognition}

\addauthor{Soumitri Chattopadhyay}{soumitri@cs.unc.edu}{1}
\addauthor{Sanket Biswas}{sbiswas@cvc.uab.es}{2}
\addauthor{Emanuele Vivoli}{evivoli@cvc.uab.es}{2, 3}
\addauthor{Josep Llad\'{o}s}{josep@cvc.uab.es}{2}

\addinstitution{
 Department of Computer Science,\\University of North Carolina at Chapel Hill, USA}
 
\addinstitution{Computer Vision Center \& Computer Science Department,\\Universitat Autònoma de Barcelona, Spain
}

\addinstitution{MICC, University of Florence, Italy}

\runninghead{Chattopadhyay et al.}{Generative Class Prompt Learning}

\begin{document}

\maketitle

\begin{abstract}

\noindent
Although foundational vision-language models (VLMs) have proven to be very successful for various semantic discrimination tasks, they still struggle to perform faithfully for fine-grained categorization. Moreover, foundational models trained on one domain do not generalize well on a different domain without fine-tuning. We attribute these to the limitations of the VLM's semantic representations and attempt to improve their fine-grained visual awareness using generative modeling. Specifically, we propose two novel methods: \textit{\textbf{G}enerative \textbf{C}lass \textbf{P}rompt \textbf{L}earning} (GCPL) and \textit{\textbf{Co}ntrastive \textbf{M}ulti-class \textbf{P}rompt \textbf{Le}arning} (CoMPLe). Utilizing text-to-image diffusion models, GCPL significantly improves the visio-linguistic synergy in class embeddings by conditioning on few-shot exemplars with learnable class prompts. CoMPLe builds on this foundation by introducing a contrastive learning component that encourages inter-class separation during the generative optimization process. Our empirical results demonstrate that such a generative class prompt learning approach substantially outperform existing methods, offering a better alternative to few shot image recognition challenges. The source code will be made available at: \url{https://github.com/soumitri2001/GCPL}.


\end{abstract}


\section{Introduction}\label{sec:intro}

Foundational vision-language models (VLMs) \cite{vlm_survey1, clip, align} have emerged as powerful open-world learners due to their large-scale web-based pre-training, covering a broad range of multi-modal concepts. Models like CLIP \cite{clip} and ALIGN \cite{align}, trained with a contrastive learning objective to align image-text pairs, show promising generalization to downstream tasks \cite{vlm_fsl_survey}, often in a zero-shot manner \cite{sain2023clip, zhou2023zegclip}. CLIP-based zero-shot classification, for example, uses a handcrafted prompt ("A photo of a \texttt{[CLASS]}") to represent a class, predicting the class by computing the similarity between text and image embeddings \cite{clip, coop}.


However, the performance of zero-shot CLIP baseline is often suboptimal, especially for fine-grained image recognition tasks \cite{flowers, cars, cornseeds}. This limitation is largely due to the quality of handcrafted prompts, which lack discriminative task-specific knowledge \cite{coop, cocoop, maple}. To mitigate this issue, techniques such as \textit{adapter learning} \cite{tip_adapter, svladapter} and \textit{prompt learning} \cite{coop, maple, prograd, kgcoop} have emerged in the VLM space. While the former introduces extra layers on top of frozen vision and language representations for alignment \cite{svladapter}, prompt learning, inspired by~\cite{nlp_prompting}, transforms the previously handcrafted prompts (e.g., "A photo of a") into soft, learnable prompts, with the \texttt{[CLASS]} token fixed. This enables the prompts to capture task-specific contextual information, optimized over few-shot support sets. Subsequent work \cite{maple, prograd, kgcoop, plot} has further refined this approach, enhancing contextual learning and improving visiolinguistic synergy \cite{maple, plot}.

\begin{figure}[t]
    \centering
    \includegraphics[width=\columnwidth]{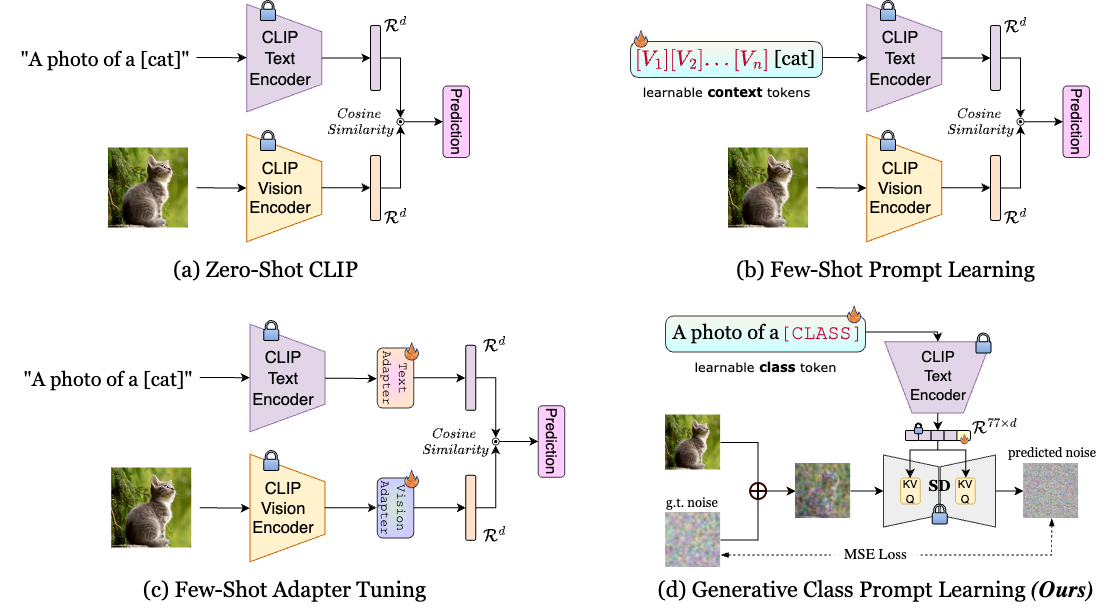}
    \vspace{-0.6cm}
    \caption{Overview of our approach compared to existing VLM adaptation methods. (a) Zero-shot inference with CLIP; (b) Contextual prompt token learning; (c) Adapter-based tuning with handcrafted prompts on frozen CLIP representations; (d) \textbf{Our setup (GCPL)}: generatively learning the \texttt{[CLASS]} token by prompting a frozen text-to-image LDM \cite{ldm}).}
    \label{fig:teaser}
\end{figure}

Despite improvements, several issues persist in CLIP's transferability to specialized downstream tasks, particularly in fine-grained classification. Firstly, fine-grained category names are often highly dataset-specific or domain-specific, lacking the visual semantic knowledge that could assist in accurate recognition. For example, consider the FGVCAircraft \cite{fgvc_aircraft} dataset that has aircraft model names like \texttt{Tu-154, DR-400, MD-80} etc., or StanfordCars \cite{cars} with fine-grained vehicular models such as \texttt{``2012 Acura TL Sedan,'' ``2011 Audi S6 Sedan''}, etc. Such customized categories \textit{do not} contain sufficient relevant visual cues and would naturally lead to spurious representations when fed through CLIP. 
Although multimodal prompt learning \cite{maple, promptsrc} seeks to enhance visio-linguistic synergy, it depends on frozen CLIP embeddings, causing their suboptimality to transfer to learnable tokens via self-attention \cite{vaswani, vit}. Secondly, CLIP's training on web-scraped data focuses on natural visual content, limiting its ability to generalize to specialized domains such as medical imaging \cite{crc, isic}, affecting multimodal prompting \cite{maple} due to a lack of domain-specific knowledge in both visual and textual branches. Thirdly, certain visual concepts are \textit{hard to describe by text} (e.g., abstract fractals \cite{fractals}), making CLIP’s class embeddings less effective. Thus, we can single out the core underlying problem: suboptimality of raw CLIP representations, which often lack fine-grained visual semantic awareness.

In this work, we challenge the existing paradigm of learning contextual prompts \cite{coop, kgcoop, maple} and instead advocate for learning stronger class embeddings with enhanced visual semantic knowledge. In our attempt to tackle these aforementioned limitations head-on, we take a radically different route from conventional approaches \cite{clip, coop, kgcoop, maple}; leveraging text-to-image diffusion models \cite{ldm} for few-shot class prompt learning. Two primary factors motivate our line of exploration: the immense potential unearthed from zero-shot generative classifiers \cite{zsdc_pathak, zsdc_iclr} and their emergent properties such as strong shape bias and human-level out-of-distribution accuracy \cite{zsdc_iclr}; and the explicit \textit{parameterization of textual conditions} for controllable generation \cite{textinv, dreambooth}. Text-to-image latent diffusion models such as Stable Diffusion (SD) \cite{ldm, controlnet} rely on a text conditional prompt that interacts with a U-Net denoiser via cross-attention \cite{ldm} during the denoising process \cite{ddpm}, creating strong visio-linguistic synergy. In our few-shot learning approach with an exemplar support set and a predefined prompt (e.g., "A photo of \texttt{[CLASS]}"), we make the \texttt{[CLASS]} token \textit{learnable} while keeping the  contextual token fixed (unlike traditional prompt tuning \cite{cocoop}). This learnable prompt embedding can be utilized as the text control for the diffusion model \cite{ldm}. The class embedding is \textit{learned} via optimisation of the mean-squared error loss between estimated and the added noises \cite{ddpm, ldm} on the exemplar images of that particular class. This approach, coined as Generative Class Prompt Learning (\textbf{GCPL}), is derived from the so-called textual inversion paradigm \cite{textinv}, with however, a \textit{different goal}: to learn rich pseudo-text-like representations that are enhanced with fine-grained visual semantics, to aid fine-grained discriminative tasks. Empirical results (Section \ref{sec:results}) highlight the high robustness and effective fine-grained feature learning achieved via GCPL as compared to existing works \cite{tip_adapter, cocoop, clip, zsdc_pathak}.


Therefore, in this way we present a very simple yet effective means to adapt a frozen SD model \cite{ldm} to learn class prompts for few-shot classification. However, a key weakness is that our method learns class embeddings directly from exemplar images without learning any inter-class discrimination, potentially leading to poor separation of latent class representations and reduced performance at a fine-grained level. To address this, we additionally propose \textbf{Co}ntrastive \textbf{M}ulti-class \textbf{P}rompt \textbf{Le}arning (\textbf{CoMPLe}), which optimizes all class prompts simultaneously using a contrastive learning approach. For each noisy image latent, the denoised output should be closest to its ground-truth noise and distinct from those of other classes. This involves optimizing two loss terms: minimizing for positive (same class) noise pairs and maximizing for negative (different class) pairs. Empirical results show that CoMPLe enhances GCPL performance, especially on out-of-domain tasks like medical imaging datasets \cite{isic, crc}, highlighting the benefits of inter-class awareness.

Summing up, our main contributions are: \textit{\textbf{(i)}} We identify potential limitations in CLIP's visio-linguistic knowledge bed with practical cases of failure; \textit{\textbf{(ii)}} we propose a generative class prompt learning baseline, leveraging pre-trained diffusion models to mitigate those limitations and advocating for learning stronger vision-induced textual representations with inter-class discriminative knowledge; \textit{\textbf{(iii)}} via thorough experimentations and comparisons with prior approaches, we demonstrate the potential and robustness of our proposal.

Being a first-of-its-kind attempt to introduce generation-aided prompting for few-shot V/L tasks, we hope our work would serve as a baseline to stir up further interest in advancing the study of generative classifiers and parameterization in diffusion models in general.

\vspace{-0.4cm}
\section{Related Work}\label{sec:lit}
\vspace{-0.25cm}

\keypoint{Few-shot Image Classification:} 
Few-shot learning (FSL) focuses on training models with a limited support set, typically formulated as an $N$-way $K$-shot classification task, where $N$ is the number of classes and $K$ is the number of examples per class \cite{fsl_survey1, closer_fsic, closerlookfsl}. Traditional FSL methods use meta-learning frameworks \cite{metabaseline, closer_fsic, closerlookfsl, rethink_fsic}, involving meta-training on base classes and meta-testing on novel classes through simulated episodes \cite{laenen2021episodes}. These methods are categorized into metric-based \cite{protonet, matchnet, relationnet}, alignment-based \cite{deepemd, parn, frn, bfrn}, and optimization-based approaches \cite{maml, maml2, reptile}. More recently, large-scale multimodal models such as CLIP \cite{clip} have enabled few-shot learning across modalities by leveraging broad world knowledge \cite{clip, kgcoop, tip_adapter}. Unlike these discriminative methods, our work introduces a \textit{generative modeling-based} framework for few-shot classification.

\vspace{-5mm}
\keypoint{Foundational Vision-Language Models (VLMs):} Using language as a supervisory signal with visual content has led to the development of robust multi-modal representations in large-scale VLMs like CLIP \cite{clip}, ALIGN \cite{align}, and Flamingo \cite{flamingo}. These models, trained on vast amounts of internet data (e.g., 400M image-text pairs for CLIP), employ contrastive learning \cite{cpc, simclr} to capture extensive open-world visio-linguistic knowledge \cite{vlm_survey1}. This makes them highly effective for zero-shot tasks such as classification \cite{siddiqui2021analyzing, tang2024data}, cross-modal retrieval \cite{sain2023clip, koley2024you}, anomaly detection \cite{zhou2023anomalyclip, li2024promptad}, and segmentation \cite{zhou2023zegclip, ding2022decoupling, aleem2024test}. However, despite their strong representations, models like CLIP struggle with fine-grained categorization \cite{coop, cars} and domain-specific tasks \cite{crc, isic}. To address this, methods such as linear probing \cite{clip, tip_adapter, lp++}, prompting \cite{coop, cocoop, maple, prograd, promptsrc, kgcoop}, and adapters \cite{tip_adapter, svladapter} have been introduced to adapt VLMs to specialized tasks using few-shot support sets, enhancing their performance \cite{vlm_peft_survey, vlm_fsl_survey}. Inspired by these adaptation techniques, we propose a \textit{generation-aided} class prompt learning method that combines CLIP \cite{clip} and Stable Diffusion \cite{ldm} to learn strong few-shot exemplar representations.

\vspace{-5mm}
\keypoint{Prompt Learning in VLMs:} Originating from NLP \cite{nlp_prompting}, prompt learning is a parameter-efficient method to adapt foundational models like CLIP \cite{clip} for downstream tasks in few-shot settings \cite{vlm_peft_survey}. The standard handcrafted prompt “a photo of a \texttt{[CLASS]}” used in zero-shot CLIP is suboptimal \cite{coop, kgcoop} as it lacks task-specific information. To address this, prompt learning replaces fixed context words with learnable soft prompts that encode task-specific knowledge. Methods like CoOp \cite{coop} and CoCoOp \cite{cocoop} optimize continuous context vectors in CLIP's text branch, while MaPLe \cite{maple} applies prompt learning to both text and vision branches for better visio-linguistic synergy. Further advances include techniques such as optimal transport \cite{plot}, self-regularization \cite{promptsrc}, and consistency regularization \cite{coprompt}. All these methods keep the \texttt{[CLASS]} embedding frozen while learning context tokens. Our approach differs by optimizing the \texttt{[CLASS]} embedding while using handcrafted context words, learning class prototypes via prompting with a frozen text-to-image diffusion model \cite{ldm} (details in Section \ref{sec:method}).

\vspace{-5mm}
\keypoint{Diffusion models for discriminative tasks:} With recent advances, diffusion models \cite{ddpm, ldm} have become a standard for visual content generation \cite{ldm, textinv, dreambooth}, relying on an iterative denoising process \cite{ddpm} that enables \textit{controllable} image generation through proper conditioning \cite{clf_free_guidance, uni_guidance, ldm, controlnet}. This same process can be adapted for discriminative tasks, such as classification \cite{zsdc_clark, zsdc_pathak, gazediff}, image retrieval \cite{koley2024text}, and segmentation \cite{peekaboo, freesegdiff}, by parameterizing and learning the conditions via the diffusion model's reconstruction objective \cite{ldm}. Many of these approaches use Stable Diffusion, similar to CLIP \cite{clip}, to leverage large-scale, open-world knowledge for zero-shot inference. Zero-shot adaptations of SD for classification \cite{zsdc_clark, zsdc_pathak, zsdc_iclr} benefit from the unique properties of generative classifiers \cite{zsdc_iclr}. However, like zero-shot CLIP, naive use of SD faces similar challenges, which we address by introducing a few-shot generative classifier that learns class-specific prototypes as conditional inputs to SD \cite{textinv, ldm}.

\section{Proposed Methodology}\label{sec:method}
\vspace{-2mm}
\keypoint{Overview:} We propose a class prototype learning framework using a pre-trained text-to-image diffusion model, using few-shot exemplar sets per class. Specifically, we derive from \cite{textinv} and parameterize a handcrafted prompt (e.g. ``A photo of a \texttt{[CLASS]}''), where the \texttt{[CLASS]} token is made \textit{learnable} keeping the rest of the prompt fixed. This prompt is then fed through CLIP's text encoder \cite{clip} to obtain a continuous text embedding, which is then used to condition the denoising process of a frozen Stable Diffusion model \cite{ldm} that aims to reconstruct the few-shot images provided for that particular class (Section \ref{sec:cpl}). This is our vanilla few-shot learning approach that learns each \texttt{[CLASS]} embedding independently; we further extend this to \textbf{CoMPLe}, a \textbf{co}ntrastive \textbf{m}ulti-class \textbf{p}rompt \textbf{le}arning setup where all the class prompts are optimized simultaneously using a contrastive reconstruction loss objective that minimizes the noise prediction error for a sample's true noise while maximizing the error for other (i.e. negative) samples (Section \ref{sec:comple}). For classification, we use Diffusion Classifier \cite{zsdc_pathak}, replacing CLIP class embeddings (zero-shot) with our \textit{few-shot learned embeddings}.

\begin{figure}[t]
    \centering
    \includegraphics[width=\columnwidth]{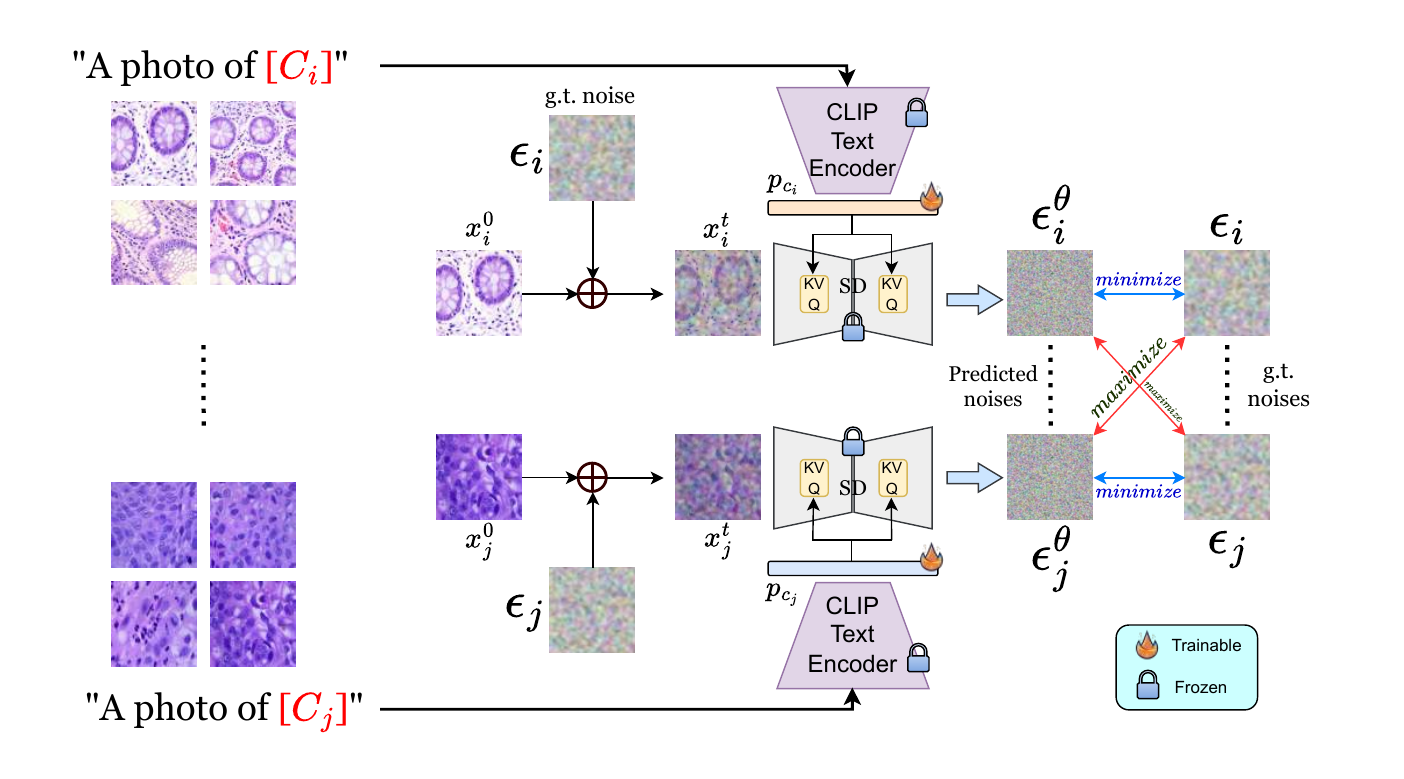}
    \vspace{-0.8cm}
    \caption{
    \textbf{Contrastive multi-class prompt learning (CoMPLe)} framework. Our proposed CoMPLe learns class prompts by optimizing the LDM loss for the trainable class token, minimizing noise reconstruction for ground truth noise while maximizing it for other class noises. Red arrows show "maximize," and blue arrows show "minimize." The diffusion classifier \cite{zsdc_pathak} uses our few-shot learned \texttt{[CLASS]} embeddings for inference.
    }
    \label{fig:mainfig}
    \vspace{-2mm}
\end{figure}

\vspace{-0.5mm}
\subsection{Revisiting Text-conditional Latent Diffusion Models}\label{sec:t2i} 


\keypoint{Preliminaries of LDM.} The working principle of diffusion models \cite{ddpm} is based on two complementary iterative processes of adding noise (forward) and removing noise (reverse) to a given clean image. Both processes are spaced over $T$ timesteps. Latent diffusion models (LDM) \cite{ldm} shifted the diffusion process i.e. noise addition and removal processes from pixel space to the latent space, projected by a pre-trained VAE encoder $\mathcal{E}(\cdot)$ and eventually decoded back to the image space by a pre-trained VAE decoder $\mathcal{D}(\cdot)$, such that $\mathcal{D}(\mathcal{E}(x))\approx x$. Formally, a clean image latent $x_0$ undergoes iterative Gaussian noise addition across $T$ timesteps, such that at a given timestep $t$ the noisy image $x_t = \sqrt{\bar{\alpha}_t}x_{0} + \sqrt{1-\bar{\alpha}_t}\epsilon$, where $\epsilon \sim \mathcal{N}(0,I)$ is the noise added, $\{\alpha_t\}^{T}_{1}$ refers to a noise schedule where $\bar{\alpha}_t = \prod^{t}_{i=1} \alpha_i$ \cite{koley2024text}. A denoising U-Net parameterizes the noise removal process \cite{unet} $\Phi_\theta$, taking the noisy latent as input and aims to estimate the added noise $\epsilon \approx \Phi_\theta(x_t, t)$, trained using a reconstruction (i.e. MSE loss). Having learned how to effectively denoise samples, the denoiser network $\Phi_\theta$ can now start from pure Gaussian noise and employing $T$ timesteps, it can produce cleaner images $\{x_{t-1}, x_{t-2}... x_0\}$, mimicking in the original distribution of $x_0$ \cite{ddpm}.

\keypoint{Text-conditioned LDM.} The unconditional denoising diffusion model \cite{ddpm} can be made ``conditional'' by fusing auxiliary control signals with the denoiser U-Net $\Phi_\theta$. The control signals may be of various modalities, e.g. textual prompts \cite{ldm, controlnet, textinv, neti}, semantic maps \cite{controlnet, dreambooth}, 2D pose \cite{controlnet} etc. We take particular focus in text-conditioned LDM, where a pre-trained text encoder like CLIP \cite{clip} is used to obtain tokenized textual embeddings from raw text prompts, which interact with the denoiser network through cross-attention (shown in \autoref{fig:mainfig}). This brings out the visio-linguistic synergical modeling in text-conditioned LDMs (in particular, we use Stable Diffusion \cite{ldm}), which we exploit in our work.

\vspace{-3mm}
\subsection{Generative Class Prompt Learning}\label{sec:cpl}

Given a few-shot support set and a corresponding prompt containing the class name, we aim to learn a visually enriched representation for the \texttt{[CLASS]} token present in our prompt embedding. Taking a leaf out of \cite{textinv}, we first tokenize the prompt to unique word embeddings \cite{clip}, and make the class-corresponding token \textit{learnable}. To do so, we introduce a placeholder token, $p_c$, into the vocabulary of CLIP \cite{clip}, and make the parameterization trainable. Now we leverage the frozen SD model \cite{ldm} to perform conditional denoising over the given set of images, with the textual condition containing the learnable token $p_c$. The optimization of SD follows the standard $l_2$ loss between the predicted and ground truth noise added to the respective latents. The optimization follows the following equation:
\vspace{-2mm}
\begin{equation}
    \mathcal{L}_{GCPL} = \mathbb{E}_{x\sim\mathcal{E}(x), p_c, \epsilon \sim \mathcal{N}(O,1), t} \left[ \lVert \epsilon_{c} - \epsilon^{\theta}_{c}(x_t, t, c_\theta(p_c)) \rVert^2_2 \right]
\end{equation}
\vspace{-4mm}

where $\epsilon_c$ is the ground truth noise added to the image latent of class `$c$', $\epsilon^{\theta}_{c}$ being the noise predicted by the denoiser parameterized by  $\theta$, and $c_\theta$ maps the learnable class prompt $p_c$ to a conditioning embedding. Essentially, $p_c$ is learned via:
\vspace{-2mm}
\begin{equation}\label{eq:ti}
    p^{*}_c = \underset{p_c}{\mathrm{\arg\min}}\text{    } \mathbb{E}_{x\sim\mathcal{E}(x), p_c, \epsilon \sim \mathcal{N}(O,1), t} \left[ \lVert \epsilon_{c} - \epsilon^{\theta}_{c}(x_t, t, c_\theta(p_c)) \rVert^2_2 \right]
\end{equation}. 
\vspace{-4mm}

\noindent This generative optimization induces a synergy between the class prompts and the visual exemplars fed into the LDM, resulting in visually enhanced class prototypes (refer to \autoref{fig:mainfig}).

\vspace{-3mm}
\subsection{CoMPLe: Contrastive Multi-Class Prompt Learning}\label{sec:comple}

As discussed previously, mere learning of class prompts, without any discriminative knowledge of other classes, can potentially lead to suboptimal class embeddings, given our primary goal is downstream discriminative tasks. For this, we extend GCPL to a multi-class setting where all class prompts $\{p_{c_i}\}$ are initialised together and simultaneously optimized via the above-discussed objective. We identify a simple contrastive constraint: in addition to aligning the predicted noises with our ground truth added ones, we also \textit{enforce divergence of predicted noises against the ground truths noises of other classes.} As shown in \autoref{fig:mainfig}, we apply two loss terms: a minimization objective over true noise pairs (i.e. belonging to same class), and a maximization objective over negative pairs of noises. Together, the loss function across a batch of $B$ samples containing $c_{\{1..B\}}$ classes can be written as: 

\vspace{-4mm}
\begin{equation}
\begin{split}
\mathcal{L}_{CoMPLe} = \frac{1}{B}\sum^B_{i=j}\mathbb{E}_{x\sim\mathcal{E}(x),p_{c_j},\epsilon \sim \mathcal{N}(O,1), t} \left[ \lVert \epsilon_{c_i} - \epsilon^{\theta}_{c_j}(x^j_t, t, c_\theta(p_{c_j})) \rVert^2_2 \right] \\ - \lambda \cdot \frac{1}{B(B-1)}\sum^B_{i\neq j}\mathbb{E}_{x\sim\mathcal{E}(x), p_{c_j}, \epsilon \sim \mathcal{N}(O,1), t} \left[ \lVert \epsilon_{c_i} - \epsilon^{\theta}_{c_j}(x^j_t, t, c_\theta(p_{c_j})) \rVert^2_2 \right]
\end{split}
\end{equation}
\vspace{-1mm}

where $\lambda$ is a weighting hyperparameter that balances the two contrasting MSE objectives. Learning of $\{p_{c_i}\}$ follows the optimization similar to \autoref{eq:ti}. This allows us to imbibe inter-class discriminative knowledge within our generatively trained class prompting setup, boosting fine-grained representation learning.

\vspace{-2mm}
\subsection{Few-shot Diffusion Classifier}\label{sec:diffcls}
\vspace{-1mm}

We adopt the recently proposed diffusion classifier \cite{zsdc_pathak} and extend it to our few-shot inference setup with minimal changes, which ensures fair comparisons and robustness.

\keypoint{Preliminaries.} As discussed in preceding sections, an LDM \cite{ldm} parameterizes the reverse process $p_\theta(x_{t-1} \mid x_t, c)$, i.e. denoising $x_t$ to $x_{t-1}$, where $c$ is the conditioning signal. Now, using $x_t = \sqrt{\bar{\alpha}_t}x_{0} + \sqrt{1-\bar{\alpha}_t}\epsilon$ (refer to Section \ref{sec:t2i}), diffusion model learns the noise estimation network $\epsilon_\theta(x_t, c)$, and hence following \cite{zsdc_pathak} the variational lower bound (ELBO) may be rewritten as $-\mathbb{E}_\epsilon \left[ \sum^T_{t=2} w_t \lVert \epsilon - \epsilon_\theta(x_t, c) \rVert^2_2 - \log p_\theta(x_0 \mid x_1, c) \right] + C$, which can be further simplified to $-\mathbb{E}_{\epsilon, t} \left[ \lVert \epsilon - \epsilon_\theta(x_t, c) \rVert^2_2 \right] + C$, assuming $w_t=1$ and $\log p_\theta(x_0 \mid x_1, c) \approx 0$ for large $T = 1000$ \cite{zsdc_pathak, gazediff}. 

\keypoint{Classification.} For classification \cite{zsdc_pathak, gazediff}, we leverage Bayes' Theorem on model predictions $p_\theta(x \mid c_i)$ over labels $\{c_i\}$ and obtain the inverse probability: $p_\theta(c_i \mid x) = \frac{p(c_i)p_\theta(x \mid c_i)}{\sum_j p(c_j)p_\theta(x\mid c_j)}$. Simplifying $p_\theta(x \mid c)$ with the ELBO derived above, we can rewrite the formulation as: 

\vspace{-2mm}
\begin{equation}\label{eq:dc}
    p_\theta(c_i \mid x) = \frac{\exp \{ -\mathbb{E}_{\epsilon, t} \left[ \lVert \epsilon - \epsilon_\theta(x_t, c_i) \rVert^2_2 \right] \}}{\sum_j \exp \{ -\mathbb{E}_{\epsilon, t} \left[ \lVert \epsilon - \epsilon_\theta(x_t, c_j) \rVert^2_2 \right] \}} 
\end{equation}
\vspace{-2mm}

Following \cite{zsdc_pathak}, an unbiased Monte Carlo estimate is calculated for each expectation by sampling $N(t_i, \epsilon_i)$ pairs and computing $\frac{1}{N}\sum^N_{i=1}\lVert \epsilon_i - \epsilon_\theta(\sqrt{\bar{\alpha}_{t_i}}x + \sqrt{1-\bar{\alpha}_{t_i}}\epsilon_i, c_j) \rVert^2_2$. Plugging this quantity in \autoref{eq:dc} gives us the diffusion classifier. 

\keypoint{Adapting to our few-shot setting.} In our framework, the conditioning signal $c$ is derived from the learned class prompts, i.e. $c_i = c_\theta(p_{c_i})$. Using this value, along with a further approximation of \autoref{eq:dc} noted by \cite{zsdc_pathak}, leads to the following equation:

\vspace{-3mm}
\begin{equation}
    p_\theta(c_i \mid x) = \frac{1}{\sum_j exp \{ \mathbb{E}_{\epsilon, t} \left[ \lVert \epsilon - \epsilon_\theta(x_t, c_\theta(p_{c_i})) \rVert^2_2 - \lVert \epsilon - \epsilon_\theta(x_t, c_\theta(p_{c_j})) \rVert^2_2 \right] \}}
\end{equation}
\vspace{-2mm}

\noindent Thus, given a test image sample $x$, we can predict its class label $c_i$ via conditional denoising across all class prompts $\{p_{c_i}\}$ and estimating the relative differences in prediction errors across each condition (i.e. $c_\theta(p_{c_i})$). Note that the class prompts $\{p_{c_i}\}$ have been learned via GCPL (or CoMPLe) in a few-shot manner, hence we call it a \textit{few-shot diffusion classifier}.

\vspace{-4mm}
\section{Experiments}

\vspace{-2mm}
\subsection{Datasets} 

We conduct experiments on six diverse image classification datasets with fine-grained categories. span across multiple domains of computer vision, making our empirical observations generalizable and robust. 

\keypoint{Natural images:} We use StanfordCars \cite{cars} comprising $196$ fine-grained car model categories, Flowers102 \cite{flowers} consisting of $102$ floral subspecies and Cornseeds \cite{cornseeds} with $4$ seed image variants. 

    
\keypoint{Medical imaging:} We use the CRC5k \cite{crc}, ISIC2018 \cite{isic} and LC25000 \cite{lc} datasets. While CRC5k and LC25000 are histopathology datasets comprising 8 colorectal and 5 lung-colon tissue categories respectively, ISIC2018 is a dermatology image corpus spanning 7 types of skin lesions. These can be considered \textit{fine-grained} image datasets, as all image categories belong to the same anatomical region and only differ in minute attributes. 


\keypoint{Hard-to-describe visual concepts:} We use the Fractals \cite{fractals} dataset, comprising $60$ categories of intricate fine-grained visual patterns. We conjecture these abstract images are hard to describe via a hard-coded word in a text vocabulary \cite{clip} (intuitively supported by empirical findings), which drives the motivation of prompt learning class prototypes using a generative approach that imbibes visual attributes into the learned embeddings.  

\vspace{-2mm}
\subsection{Implementation} 

Our model is implemented in PyTorch \cite{pytorch} accelerated by Nvidia RTX A6000 48GB GPU. We use Stable Diffusion v1.4 \cite{ldm} as the frozen text-to-image diffusion model that encapsulates a CLIP \cite{clip} text encoder with embedding dimension $d=768$. For class prompting, hand-crafted prompt templates are used that contains dataset-specific context \cite{zsdc_pathak} words (shown in \autoref{tab:prompts}) as well as the learnable class token. For GCPL, we follow \cite{textinv} and set the following training hyperparameters: AdamW \cite{adamw} optimizer ($\beta_1=0.9, \beta_2=0.999, \omega_{decay} = 0.01$) with learning rate of $5e-4$ and trained for $2000$ epochs. For CoMPLe, the learning rate was set to $1e-3$, batch size of $4$, and trained for $4000$ epochs; the contrastive weighting factor $\lambda$ being set to $0.001$ (based on empirical findings). For downstream inference, we followed the protocol outlined in \cite{zsdc_pathak}, replacing the CLIP \cite{clip} class embeddings with \textit{our prompt learned embeddings} and using them for the conditional denoising process \cite{ldm, zsdc_pathak}.

\begin{table}[tbh]
    \centering
    \resizebox{\linewidth}{!}{
    \begin{tabular}{cccc}
    \toprule
         Dataset & Visual concept & Prompt template & Initializer word\\ \midrule
         StanfordCars \cite{cars} & Vehicular variants & ``A photo of \texttt{[CLASS]}, a type of car.'' & car \\ 
         Cornseeds \cite{cornseeds} & Natural images, agriculture & ``A photo of \texttt{[CLASS]} corn seed.'' & seed \\
         \midrule
         CRC5k \cite{crc} & Histopathology & ``\texttt{[CLASS]} tissue.'' & tissue \\
         ISIC2018 \cite{isic} & Dermatology & ``\texttt{[CLASS]} skin lesion.'' & skin \\
         LC25000 \cite{lc} & Histopathology & ``\texttt{[CLASS]} tissue.'' & tissue \\
         \midrule
         Fractals \cite{fractals} & Abstract imagery & ``\texttt{[CLASS]} fractal.'' & fractal \\
    \bottomrule
    \end{tabular}}
    \vspace{-0.3cm}
    \caption{Handcrafted prompt templates used for different datasets, where \texttt{[CLASS]} denotes the learnable class prompt token.}
    \label{tab:prompts}
    \vspace{-3mm}
\end{table}

\vspace{-1mm}
\subsection{Competitors} 
\vspace{-2mm}
\keypoint{Zero-shot approaches.} First, we compare our approach against zero-shot baselines such as CLIP \cite{clip} and Diffusion Classifier \cite{zsdc_pathak}. Both of these baselines quantify the innate representational knowledge embedded in their respective foundational models \cite{clip, ldm}, with one being discriminative and the other generative. Since the results reported in \cite{zsdc_pathak} were obtained using SD v2.0 (i.e. different checkpoint), we rerun their codes using SD v1.4 and report the results.

\keypoint{Few-shot VLM adaptation methods.} As discussed earlier, two prominent directions of few-shot VLM adaptation include use of learnable adapters \cite{tip_adapter} and prompt learning \cite{maple, kgcoop, cocoop}. We compare our approach with (a) Tip-Adapter and Tip-Adapter-F \cite{tip_adapter}, the latter being a fine-tuning variant of the otherwise training-free version; and (b) SoTA few-shot prompt learning methods: CoCoOp \cite{cocoop}, KgCoOp \cite{kgcoop} MaPLe \cite{maple} and PromptSRC \cite{promptsrc}. As discussed in Section \ref{sec:lit}, these methods use the fixed class embedding initialised from CLIP and optimizes the context tokens of the prompt, which is \textit{orthogonal} to our generatively-learned approach that directly optimizes the \texttt{[CLASS]} embedding in a handcrafted prompt.

\begin{table}[t]
    \centering
    \resizebox{\linewidth}{!}{
    \begin{tabular}{cccccccc}
         Method & StanfordCars \cite{cars} & Cornseeds \cite{cornseeds} & Flowers102 \cite{flowers} & CRC5k \cite{crc} & ISIC2018 \cite{isic} & LC25000 \cite{lc} & Fractals \cite{fractals} \\ \midrule
         
         \keypoint{Zero-Shot} \\ 
         CLIP \cite{clip} & 65.56 & 18.47 & 70.73 & 21.49 & 14.43 & 25.40 & 9.25 \\
         Diffusion Classifier \cite{zsdc_pathak} & 76.77 & 17.77 & 54.21 & 24.16 & 10.41 & 17.29 & 6.25 \\ \midrule
         
         \keypoint{Adapter} \\
         Tip-Adapter \cite{tip_adapter} & $65.82 \pm 0.51$ & $34.27 \pm 3.97$ & $89.28 \pm 0.55$ & $59.90 \pm 2.18$ & $33.88 \pm 7.26$ & $80.48 \pm 1.93$ & $81.49 \pm 1.22$ \\
         Tip-Adapter-F \cite{tip_adapter} & $75.14 \pm 0.35$ & $39.61 \pm 2.88$ & $94.25 \pm 0.43$ & $71.44 \pm 2.46$ & $40.32 \pm 5.19$ & $86.02 \pm 1.59$ & $86.16 \pm 0.54$ \\ \midrule
         
         \keypoint{Prompt learning} \\
         CoCoOp \cite{cocoop} & $71.57 \pm 0.76$ & $36.56 \pm 5.42$ & $87.84 \pm 0.48$ & $60.91 \pm 2.98$ & $24.67 \pm 6.54$ & $73.86 \pm 4.19$ & $67.89 \pm 1.29$ \\
         KgCoOp \cite{kgcoop} & $78.76 \pm 0.61$ & $38.45 \pm 4.84$ & $91.97 \pm 0.44$ & $59.90 \pm 5.17$ & $29.16 \pm 6.82$ & $75.87 \pm 3.88$ & $72.84 \pm 0.93$ \\
         MaPLe \cite{maple} & $74.39 \pm 0.43$ & $34.37 \pm 15.44$ & $93.96 \pm 0.61$ & $40.56 \pm 16.12$ & $30.33 \pm 13.67$ & $71.96 \pm 5.22$ & $76.91 \pm 6.55$ \\ 
         PromptSRC \cite{promptsrc} & $83.33 \pm 0.35$ & $33.69 \pm 4.55$ & $\textbf{97.06} \pm 0.27$ & $56.45 \pm 18.28$ & $44.18 \pm 7.02$ & $77.54 \pm 1.51$ & $\textbf{93.45} \pm 0.52$ \\ \midrule
        
         \keypoint{Ours} \\
         \rowcolor{LightCyan}
         Ours-GCPL & $\textbf{88.47} \pm 0.27$ & $43.42 \pm 2.84$ & $93.45 \pm 1.39$ & $74.76 \pm $1.94 & $48.84 \pm 2.13$ & $93.44 \pm 0.78$ & $90.76 \pm 2.23$ \\ 
         \rowcolor{LightCyan}
         Ours-CoMPLe & $87.69 \pm 1.47$ & $\textbf{45.79} \pm 2.12$ & $90.73 \pm 1.05$ & $\textbf{76.36} \pm 1.82$ & $\textbf{49.27} \pm 2.59$ & $\textbf{94.83} \pm 0.28$ & $88.83 \pm 1.57$ \\ \bottomrule
    \end{tabular}}
    \vspace{-0.25cm}
    \caption{Comparison of our proposed approaches against SoTA VLM adaptation (16-shot).}
    \label{tab:16shot_main}
    \vspace{-5mm}
\end{table}

\vspace{-3mm}
\section{Performance Analysis}\label{sec:results}
\vspace{-2mm}
\subsection{Few-shot Classification}
\vspace{-1mm}

We compare against the above-mentioned methods under few-shot settings, and report the 16-shot comparisons in \autoref{tab:16shot_main}. We observe several intriguing patterns. Firstly, zero-shot models severely fail to generalise to newer domains (medical imaging or even abstract visual patterns like Fractal). This is in accordance to our discussion early-on regarding the inherent limitations of CLIP \cite{clip}. Secondly, here we also notice the weaknesses in prompt learning approaches \cite{maple, cocoop, promptsrc}, especially under domain-shift and fine-grained categories. The high standard deviation scores for prompt learning methods on medical imaging show their noisy representations, conjecturing the importance of class token learning. Overall, our generation-aided class prompting leads to favourable performances across most datasets, and GCPL despite having no inter-class discriminatory knowledge emerges as a strong feature learning paradigm. It is interesting to note however that, though CoMPLe is supposedly an improved version of GCPL, it detoriates on datasets with larger number of classes \cite{fractals, cars, flowers}, though it shows superior performance on smaller datasets \cite{crc, lc, isic}. We note this a very intriguing observation and hope future research on this direction would possibly explore this anomaly.

\begin{figure}[t]
    \centering
    \includegraphics[width=\columnwidth]{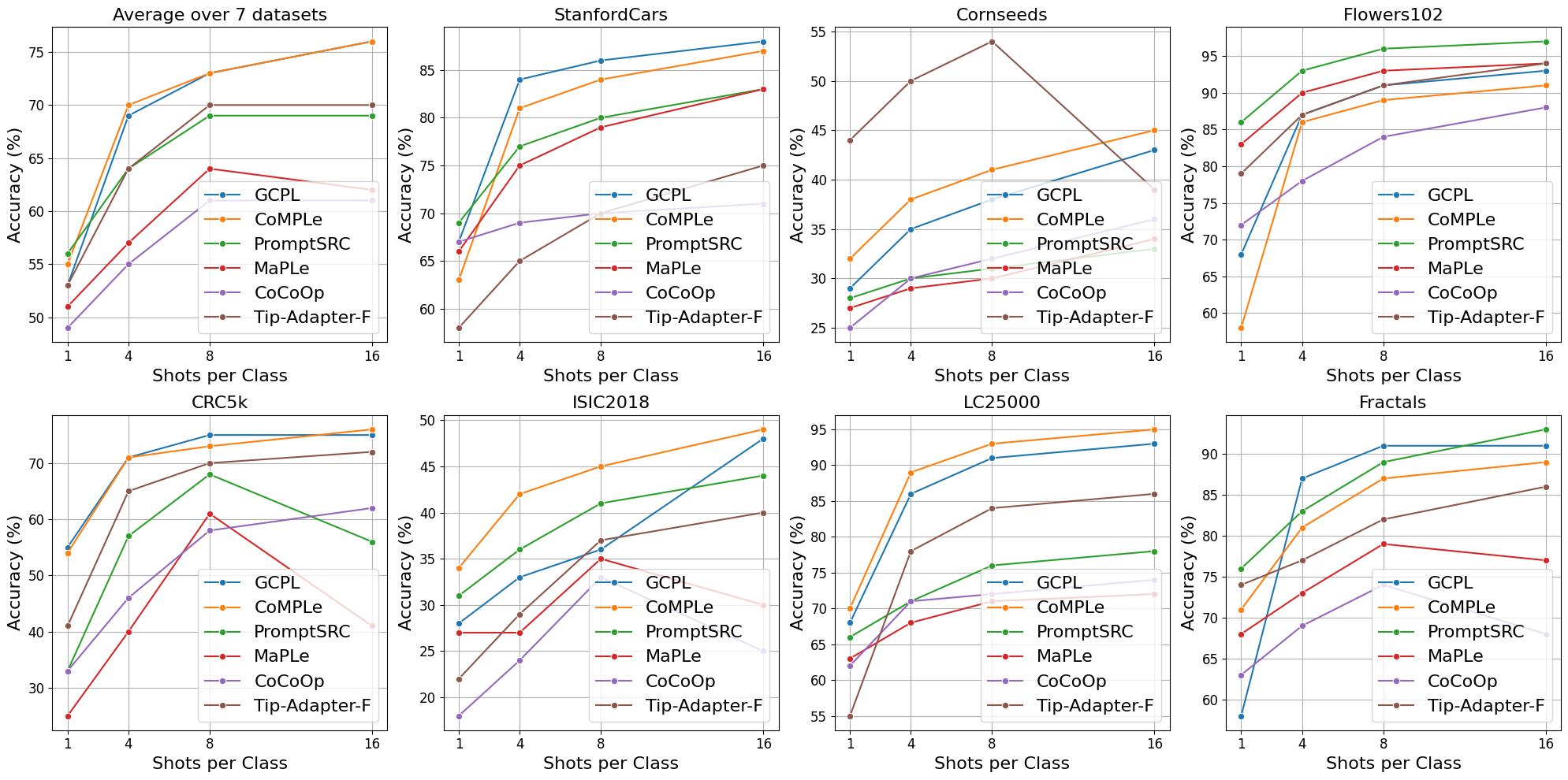}
    \vspace{-0.7cm}
    \caption{Few-shot performance across varying number of shots per class, over the 7 datasets used in this work. Note that the first sub-figure depicts the \textbf{mean} few-shot performance over all datasets (following prior works \cite{maple, cocoop}).}
    \label{fig:shots_acc}
    \vspace{-3mm}
\end{figure}

\vspace{-2mm}
\subsection{Varying the number of shots}
\vspace{-1mm}

We further experiment across different number of shots per class -- $1, 4, 8, 16$ across all of the datasets to see how our method fares in comparison to others under severely low-data settings. The results are graphically depicted in \autoref{fig:shots_acc}. We observe that both GCPL and CoMPLe show robust trends across all datasets, whereas many prompt learning approaches \cite{maple, cocoop} show spurious behavior on out-of-domain datasets. This convincingly proves that our class prompting paradigm is a strong few-shot learner.










\vspace{-1mm}
\section{Limitations}
\vspace{-2mm}
While our methods, GCPL and CoMPLe, significantly improve the fine-grained image recognition capabilities of VLMs, they also face certain limitations, which we feel are important to discuss. Diffusion models \cite{ldm}, in particular, are resource-intensive and pose a barrier to widespread adoption. GCPL's class-specific training process is inherently slow, while CoMPLe's higher GPU memory requirements, due to multiple denoising steps per batch, further increase the computational burden. This stands in contrast to prompt learning \cite{maple} and adapter tuning \cite{tip_adapter}, which are less resource-intensive. Furthermore, the need for small batch sizes in CoMPLe, driven by memory constraints, may affect performance, as larger batches typically improve contrastive learning outcomes \cite{simclr}. Moreover, the slow inference speed of the diffusion classifier, a common issue across generative classifiers \cite{zsdc_pathak, zsdc_iclr}, limits its practical deployment. Addressing these limitations will lead to more powerful and practical generation-aided discriminative models. 

\vspace{-2mm}
\section{Conclusion}
\vspace{-2mm}
In this work, we proposed a radically different paradigm for fine-grained visio-linguistic representation learning with limited data, in the form of two generative class prompt learning frameworks, GCPL and CoMPLe. We identified the potential pitfalls of CLIP representations for fine-grained and out-of-domain categories, which we were able to overcome significantly via generative modeling. Despite showing great promise, our methods require high computational resources, limiting their practical usage. Future work should focus on optimizing these methods for greater computational efficiency and scalability. As generative classifiers continue to garner interest due to their emergent properties \cite{zsdc_iclr}, we anticipate that forthcoming research will build on our work, overcoming current challenges and unlocking new capabilities in the intersection of generative and discriminative representation learning.

\vspace{2mm}
\section*{Acknowledgements}
\vspace{-1.5mm}
This work acknowledges the Spanish projects GRAIL
PID2021-126808OB-I00, DocAI 2021-SGR-01559, the
CERCA Program / Generalitat de Catalunya, and PhD Scholarship from AGAUR 2023 FI-3-
00223. The authors also acknowledge Gedas Bertasius and Feng Cheng of UNC Chapel Hill for constructive discussions and hardware resources.

\bibliography{egbib}


\end{document}